\documentclass{article}
\usepackage{roboicl,times}
\usepackage[T1]{fontenc}
\usepackage[utf8]{inputenc}
\usepackage{amsmath,amssymb,graphicx,booktabs,tabularx,longtable,array,multirow}
\usepackage{xcolor,url,xspace}
\usepackage{wrapfig}
\usepackage{hyperref}
\newcommand{\gptastra}{\texorpdfstring{\mbox{GPT-6 Astra}}{GPT-6 Astra}}
\hypersetup{colorlinks=true,linkcolor=blue!55!black,citecolor=blue!55!black,
  urlcolor=blue!55!black,pdftitle={RoboICL: Embodied In-Context Learning with GPT-6 Astra},
  pdfauthor={Fangcheng Liu, Yeqing Shen, Anda Cheng, Weishi Mi, Chao Tang, Chenyuan Liu, Yushun Xiang, Tingguang Li, Yong-Lu Li, Yehui Tang}}
\graphicspath{{figures/}}

\newcommand{\act}{\texttt{act}}
\newcommand{\train}{\textsc{Train}}
\newcommand{\live}{\textsc{Live}}
\newcommand{\R}{\mathbb{R}}
\newcolumntype{Y}{>{\raggedright\arraybackslash}X}
\AtBeginDocument{\setlength{\LTcapwidth}{\linewidth}}

\usepackage{colortbl,placeins}
\usepackage{float}
\usepackage{marvosym} 

\definecolor{resultneutral}{RGB}{245,246,248}
\definecolor{hybridfill}{RGB}{244,239,250}
\definecolor{zeroshotfill}{RGB}{235,243,252}
\definecolor{oneshotfill}{RGB}{255,243,226}
\definecolor{resultblue}{RGB}{34,67,124}
\definecolor{resultorange}{RGB}{142,79,26}
\definecolor{resultrule}{RGB}{163,174,189}

\title{RoboICL: Embodied In-Context Learning with GPT-6 Astra}

\author{%
Fangcheng Liu\textsuperscript{1\dag}, Yeqing Shen\textsuperscript{1\dag}, Anda Cheng\textsuperscript{1\dag}, Weishi Mi\textsuperscript{2}, Chao Tang\textsuperscript{2}, Chenyuan Liu\textsuperscript{3},\\
\textbf{Yushun Xiang\textsuperscript{3}, Tingguang Li\textsuperscript{2}, Yong-Lu Li\textsuperscript{3}, and Yehui Tang\textsuperscript{1\,\Letter}}\\[4pt]
\textsuperscript{1} Samsung Research, Beijing, China \qquad
\textsuperscript{2} Samsung Robotics eXperience \\
\textsuperscript{3} Shanghai Jiao Tong University\\
\texttt{\{fc.liu, yeqing.shen, anda.cheng, yehui.tang\}@samsung.com}\\[2pt]
\textsuperscript{\dag} Equal Contribution \qquad \Letter\ Corresponding Author%
}

\begin{document}
\maketitle

\begin{abstract}
General-purpose vision-language models offer a promising way to zero-shot robot control: \gptastra{} excels at open-ended and language- or image-conditioned manipulation but remains substantially weaker on high-precision and long-horizon tasks. We introduce \emph{RoboICL}, an in-context robot-control framework that narrows these gaps without robot-specific parameter updates or a learned VLA. RoboICL separates \emph{demonstration context}, which provides recorded examples when available, from \emph{interaction memory}, which accumulates the model's own actions and observed outcomes. Both use a shared observation--action--receipt--observation grammar. To preserve experience across task stages, RoboICL combines sampled demonstration blocks with bounded anchored memory. Fixed anchors keep earlier rollout interactions available for in-context learning, while the latest interaction supports immediate error correction. Across 30 RoboDojo tasks, using zero shot for Open and one demonstration elsewhere, RoboICL improves on official zero-shot \gptastra{} by 20--27 progress-score points in every category. It leads the leaderboard baselines on Memory and Open, achieves comparable performance to the strongest Precision baseline, and remains competitive on Long-Horizon. Its 30-task Overall score is 50.64, versus 33.68 for the strongest baseline. On a separate ten-task subset, RoboICL scores 60.60, within 2.00 points of the $\pi_{0.5}$ + \gptastra{} hybrid approach. On three real-robot tasks, mean progress rises from 14.45 at zero shot to 63.33 at one shot and 78.89 at three shots. On two development tasks, optional Jev-gated action reuse reduces \gptastra{} calls by 33--48\%. Code is available at \href{https://github.com/Mosi-AI/RoboICL}{https://github.com/Mosi-AI/RoboICL}.
\end{abstract}

\section{Introduction}
\label{sec:intro}
\begin{figure}[t]
  \centering
  \includegraphics[width=\linewidth]{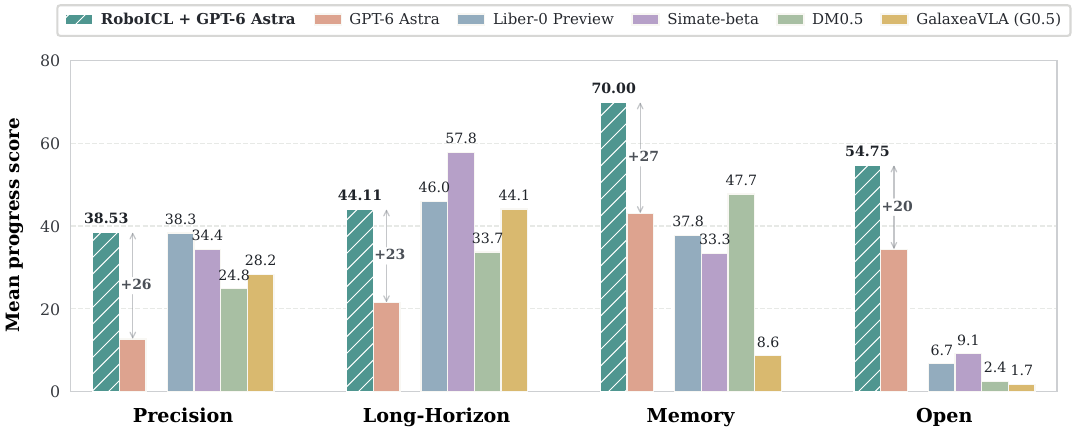}
  \vspace{-0.5cm}
  \caption[RoboDojo category-level comparison.]{\textbf{RoboDojo category-level comparison.}
  Baseline scores are reproduced from the RoboDojo leaderboard, with \gptastra{} denoting the official zero-shot results~\citep{robodojo2026astraeval}.
  As no training demonstrations are available for Open, RoboICL uses interaction memory alone (zero-shot) in this category; all other categories use
  one demonstration (one-shot). Integer labels on the official \gptastra{} bars show RoboICL's absolute gains in progress-score points. Among the strongest methods currently listed on the leaderboard, RoboICL leads on Memory and Open, is comparable to the strongest method on Precision, and remains competitive on Long-Horizon, without robot-specific post-training. Table~\ref{tab:category-comparison} reports the per-task evaluation counts.}
  \label{fig:teaser}
\end{figure}

Large language models (LLMs) can infer task patterns from examples provided at inference time, without updating their parameters~\citep{brown2020fewshot}. Extending this ability to robot control requires connecting language, images, and proprioception to continuous actions. Execution also changes the information available for the next decision: a grasp may fail, contact may stop motion, or the robot may reach a state absent from the demonstrations. Unlike language modeling, embodied control still lacks a broadly applicable foundation model that can be deployed across tasks without robotics-specific adaptation. Many Vision-Language-Action (VLA) and World-Action-Model (WAM) systems acquire their capabilities through substantial robot-specific training or post-training~\citep{fu2024icrt,vosylius2024instant,sridhar2025ricl,zhou2026zerowam,generalist2026gen15}. Recent evaluations of \gptastra{} point to a complementary possibility: a general-purpose Vision-Language Model (VLM) can generate robot actions directly~\citep{chen2026robodojo,robodojo2026astraeval}. It performs extraordinarily on open-ended and language- or image-conditioned manipulation, yet remains substantially weaker on high-precision and long-horizon tasks (Figure~\ref{fig:teaser}). The practical question is therefore how to organize context so that a fixed model can make full use of the information provided by prior examples and the experience it accumulates during execution.

Prior work provides inference-time context through several routes. Demonstration-conditioned robot policies consume sensorimotor examples~\citep{duan2017oneshot,fu2024icrt,vosylius2024instant,sridhar2025ricl}, while frozen language and vision-language models use prompted skills, programs, spatial representations, or numerical actions~\citep{dipalo2024kat,yin2025roboprompt,cheng2026gptpolicy}. Concurrent systems further combine demonstrations with execution feedback~\citep{chen2026showharness,cheng2026gptpolicy}. These advances make feedback-conditioned control feasible, while leaving open how a general-purpose VLM should organize actions, their realized consequences, and the experience accumulated over long-horizon execution.

We introduce \emph{RoboICL}, an in-context learning framework tailored to robot control. RoboICL turns inference-time context into a structured interface with two components that serve distinct roles. \emph{Demonstration context} remains fixed during an episode and presents recorded trajectories as few-shot examples. \emph{Interaction memory} updates as the robot acts and records the corresponding evidence from its own rollout, including progress, errors, and corrections. Rather than encode these sources through separate interfaces, RoboICL gives them a shared interaction grammar---observation, action chunk, execution receipt, and result observation. In demonstration context, the receipt describes the retained expert-action sequence between the two observations; in interaction memory, it reports the execution of the model-generated action chunk. The model therefore encounters the same before--action--after pattern in both sources.

RoboICL uses \emph{bounded anchored memory} to preserve the structure of long trajectories without serializing every frame. From each demonstration, it retains non-overlapping action blocks sampled across the trajectory, giving the model examples from the early, intermediate, and late stages of the task. During execution, interaction memory retains the first interaction, selected later interactions at predefined step intervals, and the latest completed interaction. The model can thus refer back to the initial scene and earlier action outcomes while using the latest feedback to decide what to do next. Whenever discarded steps separate two retained blocks, RoboICL inserts an explicit \texttt{<TRAJECTORY\_GAP>} record so that their observations are not read as consecutive.

RoboICL achieves substantially higher progress scores than official zero-shot \gptastra{} in all four categories shown in Figure~\ref{fig:teaser}. Among the compared methods, it leads on Memory and Open, is comparable to the strongest method on Precision, and remains competitive on Long-Horizon. Its 30-task Overall score is 50.64, exceeding the strongest baseline's 33.68 by 16.96 points. Open is especially notable because no demonstrations are available and RoboICL uses interaction memory alone. Separately, Table~\ref{tab:ten-task} reports a ten-task panel with 5 requested layouts per task, whereas the official RoboDojo zero-shot evaluation uses 50 episodes per task~\citep{robodojo2026astraeval}. On this smaller panel, RoboICL with \gptastra{} alone (60.60) approaches the reported performance (62.60) of GPT-as-Policy's $\pi_{0.5}$ + \gptastra{} ensemble~\citep{su2026astra}. On three single-arm real-robot tasks, mean progress score rises from 14.45 without demonstrations to 78.89 with three demonstrations. We further measure token usage and wall time, connect adaptive action horizons to model-call frequency, and study the potential of Jev-gated action reuse~\citep{jevai2026docs} to reduce inference cost (Section~\ref{sec:efficiency}). RoboICL makes three primary contributions:
\begin{itemize}
  \item \textbf{A robot-native ICL framework.} Demonstration context and interaction memory present prior examples and rollout experience to a general-purpose VLM through a shared interaction grammar, without robot-specific parameter-update. 
  \item \textbf{Minimal, auditable control.} A direct API harness exposes one function tool for continuous action chunks and uses neither an external skill library nor a general-purpose coding-agent runtime. Its model-facing loop consists of context serialization, a single action schema, and deterministic validation and execution.
  \item \textbf{Closing \gptastra{}'s gaps in high-precision and long-horizon control.} Without robot-specific post-training, RoboICL leads the compared RoboDojo leaderboard methods on Memory, Open, and 30-task Overall, while achieving comparable performance to sota baselines on Precision. The framework is further validated on three single-arm real-robot tasks.
\end{itemize}

\section{Related work}
\label{sec:related}

\paragraph{Robot learning from context.}
Few-shot language modeling established examples as an inference-time interface~\citep{brown2020fewshot}, later extended to multimodal inputs by Frozen~\citep{tsimpoukelli2021frozen} and Flamingo~\citep{alayrac2022flamingo}. In robotics, ICRT~\citep{fu2024icrt} and Instant Policy~\citep{vosylius2024instant} learn context-conditioned control from sensorimotor sequences or demonstration pairs; RICL~\citep{sridhar2025ricl} augments a pretrained VLA with demonstration retrieval; and Zero-WAM~\citep{zhou2026zerowam} and GEN-1.5~\citep{generalist2026gen15} acquire deployment-time adaptation through large-scale video, trajectory, or physical-interaction pretraining. RoboICL studies a complementary setting: it elicits action chunks from a frozen general-purpose VLM through demonstration context and interaction memory, presented in a shared interaction grammar, without training a robot-specific policy or action head.

\paragraph{Frozen models and interaction feedback.}
Frozen LLMs and VLMs control robots through skill selection~\citep{ahn2022saycan}, program generation~\citep{liang2023code}, spatial planning~\citep{huang2023voxposer}, or numerical actions from structured spatial representations~\citep{dipalo2024kat,yin2025roboprompt}. Show-Harness~\citep{chen2026showharness} uses semantic action interfaces for closed-loop control. ReAct~\citep{yao2023react} and Reflexion~\citep{shinn2023reflexion} use feedback and memory for sequential decision making, while Zeva~\citep{chen2026zeva} and Reflective VLA~\citep{lian2026reflective} study interaction histories in robotics. Among close concurrent works, GPT-as-Policy Direct~\citep{su2026astra} uses a persistent VLM agent to generate absolute end-effector targets from observations and online history without expert demonstrations or a learned VLA. GPT-Policy~\citep{cheng2026gptpolicy} additionally conditions on demonstrations and selects Cartesian waypoint and gripper tools. RoboICL instead generates per-step Cartesian-increment and gripper sequences, and uses bounded anchored memory to retain visual transitions in a shared demonstration--interaction grammar. The distinction lies in action and context representation, rather than the presence of online feedback.

\section{RoboICL: Context and Control Interface}
\label{sec:method}

\subsection{Framework overview}
\begin{figure}[t]
  \centering
  \includegraphics[width=\linewidth]{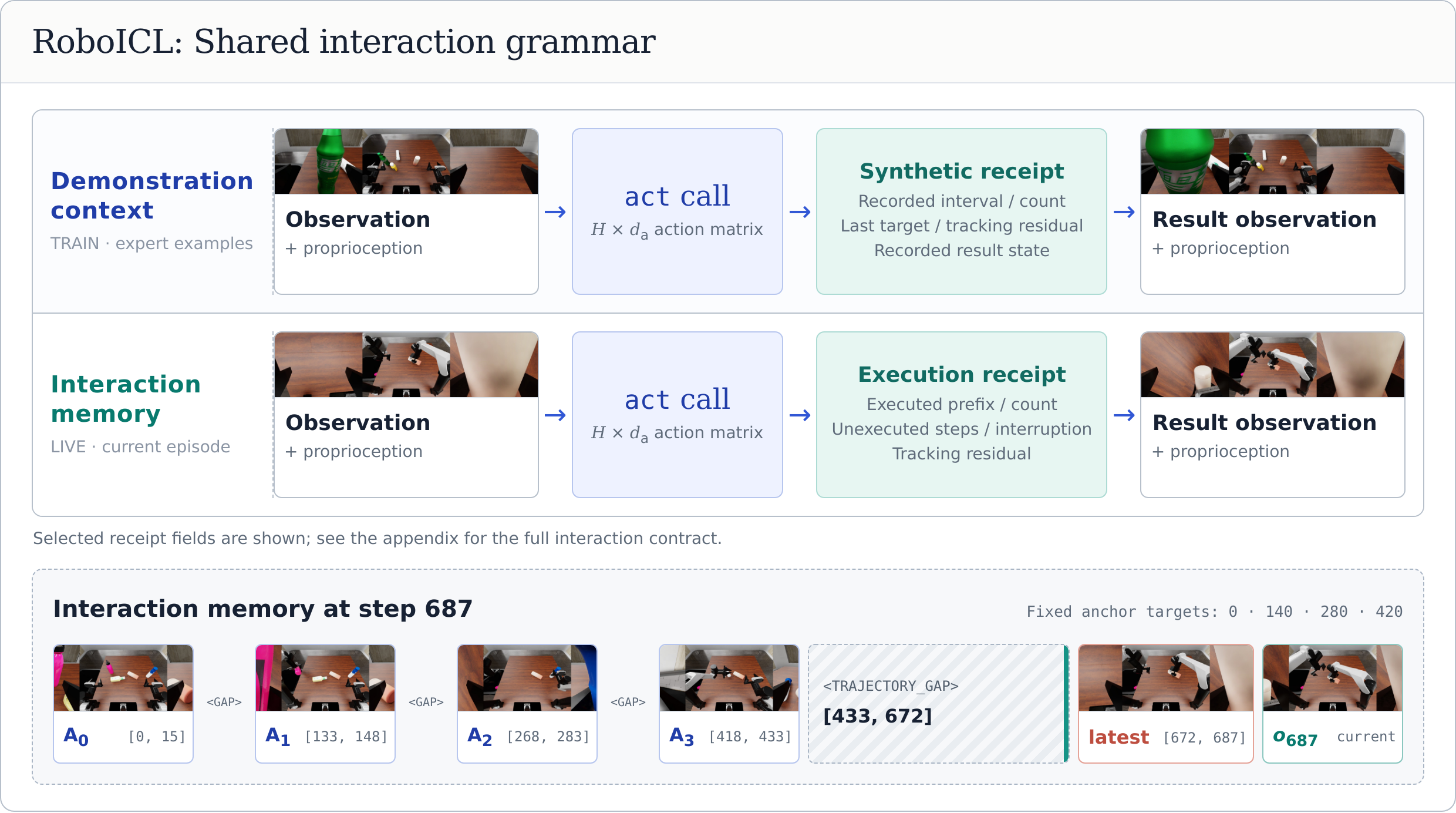}
  \vspace{-0.5cm}
   \caption{\textbf{A shared interaction grammar for demonstration context and interaction memory.} Both sources use observation--action--receipt--observation records. Synthetic demonstration receipts describe recorded intervals, action counts, target tracking, and result states. Execution receipts report realized actions and controller feedback. The general action dimension is $d_a$, with $d_a=14$ for the illustrated RoboDojo interface. The lower panel shows a bottle-task request at step 687: four fixed anchors and the latest interaction, with explicit gaps between retained intervals.}
  \label{fig:method}
\end{figure}

RoboICL organizes robot control as a sequence of model calls and executed actions. Its \emph{demonstration context} supplies recorded examples, while its \emph{interaction memory} retains experience from the current episode. Let $t$ index model decision rounds, $g$ denote the task instruction, and $(o_t,z_t)$ denote the current visual observation and proprioceptive state. Demonstration context $\mathcal D_K=\{D^{(k)}\}_{k=1}^{K}$ contains $K$ expert demonstrations and remains fixed throughout the episode; interaction memory $\mathcal M_t$ updates after execution. The serializer labels these sources \train{} and \live{}, respectively; \train{} denotes examples in the request, not parameter training. Zero-shot control uses $K=0$. With action dimension $d_a$ and prediction horizon $H$, a fixed multimodal model $\pi_\theta$ receives request $Q_t$ and predicts an action chunk $A_t$:
\begin{equation}
 Q_t=\operatorname{Serialize}(g,\mathcal D_K,\mathcal M_t,o_t,z_t),
 \qquad A_t\sim\pi_\theta(\,\cdot\mid Q_t),\quad A_t\in\R^{H\times d_a}.
 \label{eq:policy}
\end{equation}
The parameters $\theta$ remain fixed: actions are conditioned on the instruction, current state, and retained context without deployment-time gradient updates. The robot control harness calls the multimodal API directly and exposes a single \act{} function-call tool that accepts a continuous action chunk and a concise execution note. It requires neither an external skill library nor a general-purpose coding-agent runtime. Figure~\ref{fig:method} summarizes the shared interaction grammar and memory design.


The action dimension and camera configuration depend on the embodiment. In the bimanual RoboDojo, $d_a=14$: each row of $A_t$, for $i\in\{1,\ldots,H\}$, specifies seven values per arm:
\begin{equation}
 a_{t,i}=[\Delta p^L_{t,i},\Delta r^L_{t,i},u^L_{t,i},
           \Delta p^R_{t,i},\Delta r^R_{t,i},u^R_{t,i}],
 \quad \Delta p,\Delta r\in\R^3,\quad u\in[0,1].
 \label{eq:action}
\end{equation}
Here, $L$ and $R$ identify the arms, $\Delta p$ and $\Delta r$ are world-frame translation and rotation-vector increments, and $u$ is an absolute normalized gripper target ($1$ open, $0$ closed). The controlled reference point is each arm's link-6 origin, distinct from the fingertip midpoint or a calibrated tool center point (TCP); the increments remain expressed in world-aligned axes. Deterministic code validates the shape and bounds, solves inverse kinematics, and executes the admissible prefix without supplying learned action proposals. At RoboDojo's 25\,Hz control rate, a complete chunk represents $H/25$ seconds of simulated actuation, separate from model inference latency. Section~\ref{sec:efficient-interface} describes horizon selection; Section~\ref{sec:realrobot} reports the separate single-arm, two-camera real-robot study; Section~\ref{sec:jev} studies an optional gate for reusing a predicted suffix. 

\begin{figure}[t]
  \centering
  \includegraphics[width=\linewidth]{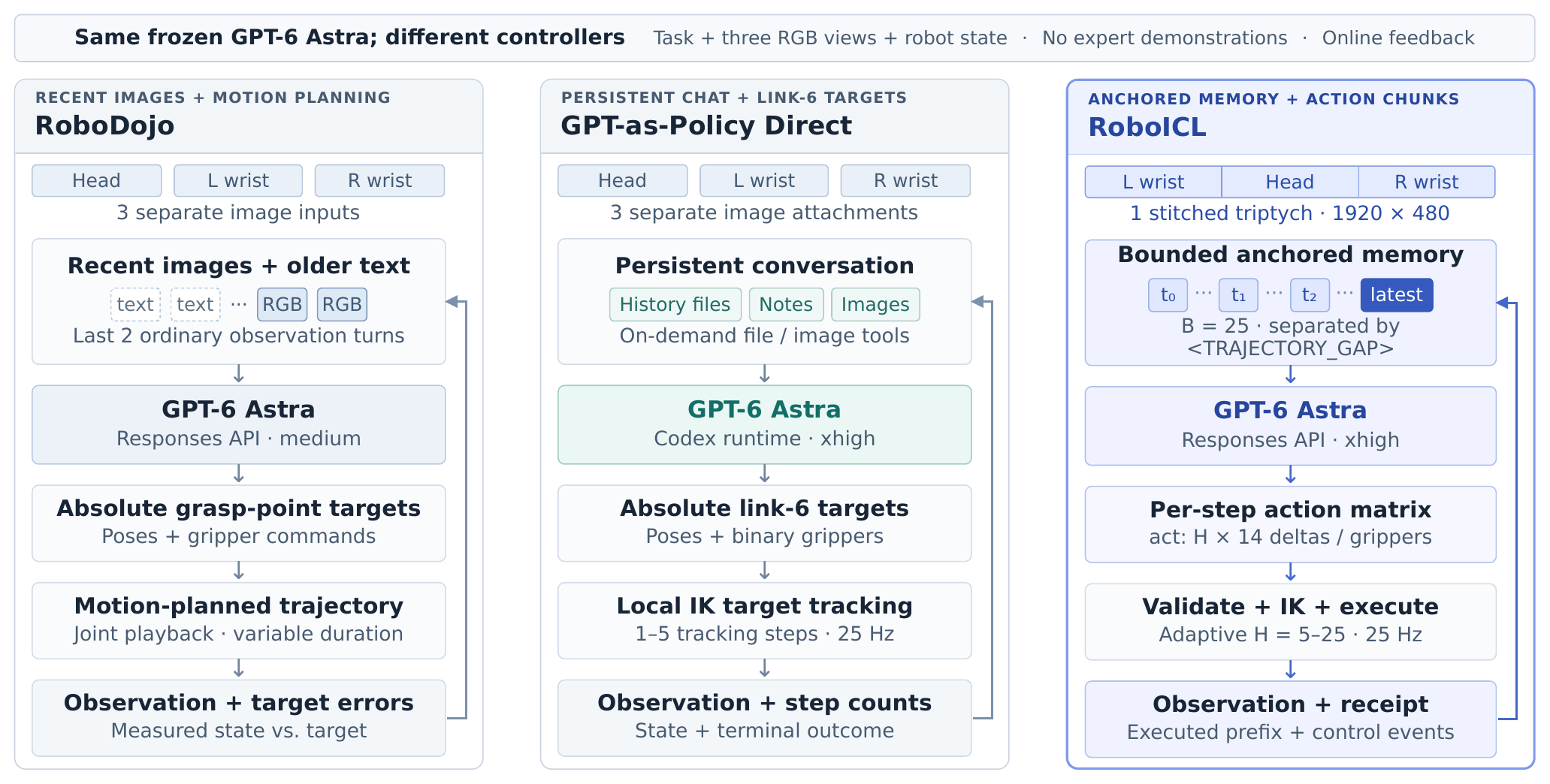}
  \vspace{-0.5cm}
  \caption[Same frozen GPT-6 Astra under three zero-shot controller designs.]{\textbf{Same frozen GPT-6 Astra under three zero-shot controller designs.} All three systems receive the task, robot state, head and bilateral-wrist images, and online feedback, without expert demonstrations or a learned VLA. The official RoboDojo controller~\citep{robodojo2026astraeval} retains recent images and older text, predicts absolute grasp-point poses and gripper commands, and motion-plans a joint trajectory lasting up to 10\,s. GPT-as-Policy Direct~\citep{su2026astra} maintains a persistent tool-accessible conversation, predicts absolute dual-arm link-6 targets, and tracks them with local inverse kinematics for 1--5 steps at 25\,Hz. RoboICL stitches the three views into one triptych, retains $B=25$ temporally distributed interactions in bounded anchored memory, and predicts an adaptive $H\times14$ sequence of per-step arm increments and gripper targets for validated execution at 25\,Hz. The comparison summarizes how context retention, image packaging, action parameterization, and execution horizon distinguish the three zero-shot controllers.}
  \label{fig:zero-shot-protocols}
\end{figure}

\subsection{Shared interaction grammar}
\label{sec:grammar}
Each request places the task instruction and demonstration context first, followed by interaction memory and the current observation and robot state. Both sources use the same interaction grammar. An interaction indexed by $j$ has the record
\begin{equation}
 c_j=\bigl((o_j,z_j),A_j,r_j,(o_{j+1},z_{j+1})\bigr),
 \label{eq:chunk}
\end{equation}
where $(o_j,z_j)$ and $(o_{j+1},z_{j+1})$ are the states before and after the interaction. In interaction memory, $A_j$ is the proposed action chunk, and $r_j$ reports predicted, dispatched, and executed step counts, the executed interval, target tracking, and controller interruptions. These counts identify any unexecuted suffix of $A_j$. In demonstration context, $r_j$ is a \emph{synthetic receipt} constructed from the recorded actions and endpoint states. It contains the source interval, action counts and indices, an action-ledger hash, the last target, its tracking residual relative to the recorded result state, and the sampled endpoint indices. All recorded actions in the selected block are represented as executed; the receipt is not measured deployment-time controller feedback. The shared grammar makes both sources readable through the same tool-call interface while preserving their provenance.

Only positive-length executed transitions are committed to anchored memory. A rejection that executes no commands remains in the immediate request tail, allowing correction from the same state without allocating another anchor. Pending correction calls and receipts are serialized with the next positive-length transition; a retained chunk can therefore contain several correction exchanges in addition to the transition in Equation~\eqref{eq:chunk}.

RoboDojo observations concatenate synchronized left-wrist, head, and right-wrist RGB views into a $1920\times480$ triptych. Proprioception includes joint states, gripper command states, and end-effector poses; the request also provides head-camera calibration and the public task description and scoring rubric. Camera geometry, action bounds, inverse kinematics, and validation are handled outside the learned policy. Implementation details are provided in the \href{https://github.com/Mosi-AI/RoboICL}{RoboICL codebase}.

\subsection{Bounded anchored memory}
\label{sec:memory}
\label{sec:prefix}

After decision round $t$, execution receipt $r_t$ and the next observation $(o_{t+1},z_{t+1})$ complete interaction $c_t$. Interaction memory incorporates this evidence through
\begin{equation}
 \mathcal M_{t+1}=\operatorname{Update}(\mathcal M_t,c_t),
 \label{eq:online-update}
\end{equation}
where the deterministic update operator applies a bounded retention policy. Let $T$ be the episode's maximum environment steps and $B$ memory slots, the first $B-1$ retain fixed temporal anchors with targets $\tau_i=\lfloor iT/B\rfloor$ for $i=0,\ldots,B-2$, and the final slot retains the latest interaction. This bounds detailed visual history independently of episode length. Explicit \texttt{<TRAJECTORY\_GAP>} records omitted intervals so that separated observations do not appear adjacent.

Demonstration context shares the same constructing method. RoboICL selects $J$ non-overlapping interaction chunks per demonstration, retaining each block's start observation, recorded expert actions, synthetic receipt, and result observation in the format of Equation~\eqref{eq:chunk}. The deterministic preparation procedure uses approximately uniform temporal targets, adjusted to action-valid windows while preserving trajectory endpoints. Gap markers identify omitted segments. Demonstration blocks provide procedural examples, while memory blocks expose rollout-specific progress and errors. Appendix~\ref{app:context-construction} specifies the selection rules and terminal-side exceptions.

The anchor design follows this in-context learning view: both demonstration context (TRAIN) and interaction memory (LIVE) retain complete interactions distributed over the trajectory, making earlier task stages available alongside recent feedback. A recent-only window preserves the same interaction grammar but progressively drops those earlier visual transitions; the latest-interaction slot preserves immediate feedback within our distributed selection.

The evaluated API configuration limits each request to 50 images. Retained interaction endpoints contribute at most $2KJ+2B$ triptychs before deduplication; separately retained terminal observations add to this count. Shared endpoints are emitted once, and a pre-inference budget check reserves capacity for the full configured interaction memory. Thus demonstration coverage and interaction memory must be allocated jointly. Section~\ref{sec:ablation} compares allocations and specifies the main zero- and one-shot budgets, including terminal-observation exceptions. The three-shot scaling study uses $K=3$, $J=5$, and $B=5$ (40 endpoints). Gap metadata and other serialized text fall outside this image bound; total tokens and inference cost vary with episode length and replanning.

\subsection{Efficiency-aware context and execution}
\label{sec:efficient-interface}
\paragraph{Stable prefixes for cache reuse.}
Bounded memory is also organized to avoid rewriting reusable context. The task instruction, demonstration messages, and fixed camera calibration form a fixed prefix. Within interaction memory, completed anchors and the closed gaps preceding them remain unchanged as the episode advances. The latest non-anchor interaction, the still-growing gap before it, and pending corrections form a mutable suffix. The request view is built separately from the stored observations and exchanges, so pruning does not modify the archived experience. Demonstration and live observations use the same image-encoding path, preserving unchanged image bytes across requests. Where supported by the model endpoint, cache-boundary hints are placed on text delimiters after complete observations or closed gaps within the stable prefix; mutable suffix content is left unmarked. This keeps completed context eligible for cache reuse as the episode advances. Section~\ref{sec:efficiency} measures realized cache reuse from provider-reported token counters, and Appendix~\ref{app:cache-verification} verifies stable request prefixes.

\paragraph{Step-limit-based action horizons.}
Action chunks amortize one model inference over several robot steps, while shorter chunks admit more frequent visual correction. We assign the RoboDojo prediction and execution horizon from the public task step limit $T$:
\begin{equation}
H(T)=
\begin{cases}
5, & T\le400,\\
10, & 400<T\le700,\\
15, & 700<T\le1100,\\
20, & 1100<T\le1600,\\
25, & T>1600.
\end{cases}
\label{eq:adaptive-horizon}
\end{equation}
The mapping is fixed within each episode and shared across the main zero- and one-shot conditions. With full execution and no early termination, the nominal call count is $\lceil T/H\rceil$; rejections and interrupted chunks can require additional calls. The design therefore trades inference frequency against feedback delay through one shared rule rather than a separate horizon for each layout or shot count. The main benchmarks use this controller without the Jev gate.

\section{Experiments}
\label{sec:experiments}
We first establish a common context configuration, then evaluate simulation performance from a ten-task panel to broader category coverage. We next test demonstration conditioning on a real robot, measure inference cost and cache reuse, and reduce model calls through Jev-gated action reuse.

\subsection{Evaluation protocol and context configuration}
\label{sec:ablation}
We evaluate simulated bimanual manipulation on RoboDojo~\citep{chen2026robodojo}. Throughout the paper, \emph{progress score} denotes normalized task credit on a 0--100 scale; its mean is distinct from the fraction of episodes that complete the task. RoboDojo uses its native scoring rules, while the real-robot study assigns credit to predefined stages. Layouts are evaluated at seed zero; Overall is an unweighted mean of task means. Each task uses the step-limit-based horizon in Equation~\eqref{eq:adaptive-horizon}, fixed within an episode and shared between the main zero- and one-shot conditions.

\begin{figure}[t]
\centering
\includegraphics[width=\linewidth]{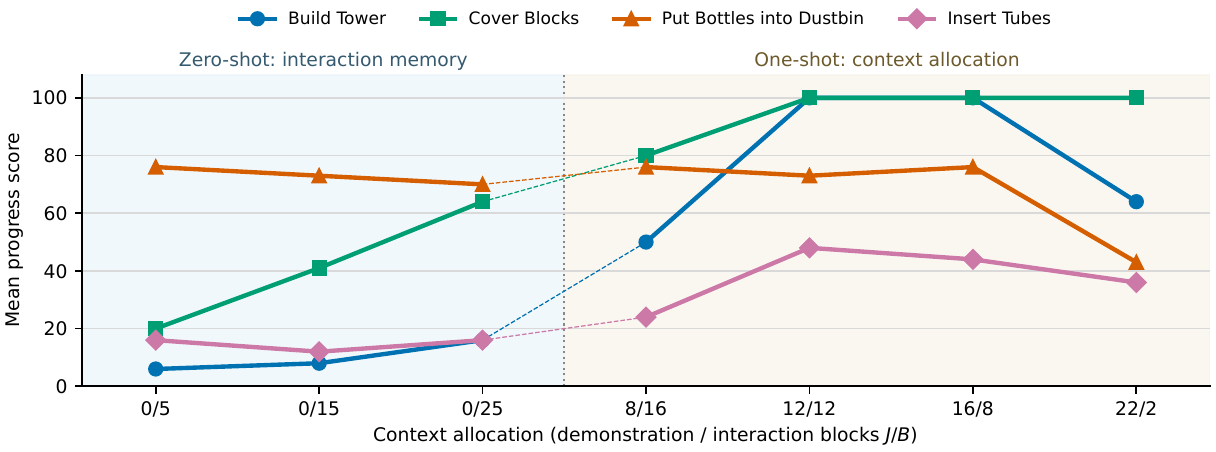}
\vspace{-6pt}
\caption{\textbf{Scaling interaction memory and allocating demonstration context.} Each point is a task's mean progress score over five layouts. The horizontal axis gives $J/B$: retained demonstration blocks and interaction-memory blocks. \textbf{Left (blue):} without demonstrations ($J=0$), increasing $B$ retains more of the robot's own experience. The plotted scores rise for Build Tower and Cover Blocks. \textbf{Right (beige):} one demonstration is sampled at different densities while $J+B=24$. The intermediate allocations $(12,12)$ and $(16,8)$ achieve the highest four-task means; allocating 22 blocks to the demonstration and only two to interaction memory lowers scores on three tasks. Thus, denser demonstration coverage alone does not guarantee better control. Dashed links connect the zero- and one-shot configurations, where both demonstration content and memory capacity change.}
\label{fig:context-allocation}
\end{figure}

\paragraph{Choosing a shared context allocation.}
With $K$ demonstrations, $J$ blocks per demonstration, and $B$ memory blocks, the total retained interaction budget is $C=KJ+B$. The one-shot study fixes $C=24$ and compares $(J,B)\in\{(8,16),(12,12),(16,8),(22,2)\}$ on four tasks with five layouts each. Figure~\ref{fig:context-allocation} gives four-task mean progress scores of 57.50, 80.25, 80.00, and 60.75, respectively. We select the balanced $J=B=12$ allocation because it achieves the highest aggregate score and use it for all main one-shot results. Build Tower and Insert Tubes also show why interaction memory cannot be reduced to two slots simply to retain denser demonstrations.

\paragraph{Anchored versus recent interaction memory.}
At the selected one-shot allocation ($K=1$, $J=B=12$), we compare anchored memory with a first-plus-recent selector that retains the first complete interaction and the latest 11 complete interactions. Demonstration blocks, all other controller settings, and the image cap are unchanged. On the same four tasks and five layouts per task used in the context-allocation study, the first-plus-recent variant achieves a mean progress score of 73.50, compared with 80.25 for the previously evaluated anchored setting. The respective gains from anchored memory are 14 points on Build Tower, 9 on Put Bottles into Dustbin, and 4 on Insert Tubes; both selectors score 100 on Cover Blocks. This controlled comparison supports the benefit of retaining temporally distributed interactions beyond the initial interaction and recent history under the evaluated one-shot configuration. Because these tasks and layouts also informed context allocation, validation on held-out tasks and layouts remains necessary to establish generality.

\paragraph{Main zero- and one-shot settings.}
The main one-shot setting uses one task-specific demonstration with $J=B=12$, yielding at most 48 retained endpoint images before deduplication. Play Tic-Tac-Toe and Play Stacking Toy additionally retain a terminal demonstration observation, for at most 49 images. Zero-shot control assigns the budget to interaction memory alone ($K=0$, $B=25$); Open and Classify Objects by Language use this setting because task-specific demonstrations are unavailable. Figure~\ref{fig:context-allocation} also reports the $B\in\{5,15,25\}$ zero-shot study. These defaults remain fixed across the following benchmark panels.

\subsection{Simulation benchmark results}
\label{sec:ten-task}
We first compare ten tasks with five requested layouts against GPT-as-Policy Direct, then expand to the four RoboDojo categories shared with the official leaderboard.

\paragraph{Zero-shot baselines.}
\label{sec:zero-shot-settings}
Zero-shot denotes the absence of supplied expert demonstrations, not the absence of task descriptions, online history, or execution feedback. Figure~\ref{fig:zero-shot-protocols} distinguishes the official RoboDojo baseline in Figure~\ref{fig:teaser}, GPT-as-Policy Direct in Table~\ref{tab:ten-task}, and zero-shot RoboICL. All three use a frozen \gptastra{} without a learned VLA, but differ in their context and control interfaces.

The official controller plans joint trajectories to absolute grasp-point targets, with distance-dependent durations of at most 10\,s and associated gripper commands after arm arrival. Direct instead uses local inverse kinematics to track absolute dual-arm link-6 targets for 1--5 control steps at 25\,Hz; RoboICL executes a validated prefix of its per-step action matrix with task-adaptive $H\in\{5,10,15,20,25\}$. All three observe the head and bilateral wrist cameras. The official controller sends three separate image content items. The released Direct runner likewise attaches three separate camera previews, each resized to a maximum edge of 480 pixels by default, while keeping originals accessible through tools. RoboICL concatenates three native $640\times480$ views into one $1920\times480$ triptych per observation; each retained endpoint therefore occupies one image slot while preserving all three views. RoboICL deterministically selects temporally distributed observation--action--receipt--observation records for every request. Appendix~\ref{app:concurrent-interfaces} compares action tools and history retention in concurrent work.

\paragraph{Comparison with GPT-as-Policy.}
Direct and RoboICL each request five layouts per task on the ten-task panel in Table~\ref{tab:ten-task}, compared with 50 episodes per task in the official evaluation. We aggregate both methods by equally averaging task means. Table~\ref{tab:ten-task} reports 60.60 for the main setting versus 45.60 for zero-shot RoboICL and 38.50 for GPT-as-Policy Direct, a 22.10-point improvement over Direct. The main setting is higher than Direct on six tasks, lower on three, and tied on one; it is within 2.00 points of the hybrid VLA--VLM mean of 62.60. Zero-shot RoboICL also exceeds Direct by 7.10 points. The largest increase from zero to one shot is on Build Tower, from 16 to 100. Imitate Sorting Sequence remains at 90, and Put Bottles into Dustbin changes only from 70 to 73, while Classify Objects drops from 100 to 71.

\begin{table}[t]
\caption{\textbf{Ten-task RoboDojo comparison.} Mean progress score (0--100); Overall averages the ten task means. Direct denotes GPT-as-Policy's Codex-based controller~\citep{su2026astra}; Figure~\ref{fig:zero-shot-protocols} compares it with the official baseline and RoboICL. RoboICL uses task-dependent horizons, with $B=25$ at zero shot and $J=B=12$ at one shot. External columns retain their published protocols~\citep{robodojo2026astraeval,su2026astra}. RoboICL targets five layouts per task and averages valid scores; $\dagger$ marks reused zero-shot results.}
\label{tab:ten-task}
\centering
\begingroup
\scriptsize
\setlength{\tabcolsep}{1.9pt}
\renewcommand{\arraystretch}{1.32}
\setlength{\aboverulesep}{0pt}
\setlength{\belowrulesep}{0pt}
\arrayrulecolor{resultrule}

\newcolumntype{N}{>{\centering\arraybackslash}X}
\newcolumntype{H}{>{\columncolor{hybridfill}\centering\arraybackslash}X}
\newcolumntype{Z}{>{\columncolor{zeroshotfill}\centering\arraybackslash}X}
\newcolumntype{S}{>{\columncolor{oneshotfill}[\tabcolsep][0pt]\centering\arraybackslash}X}
\newcommand{\resulthead}[2]{\shortstack{#1\\#2}}
\newcommand{\ourszero}[1]{\textcolor{resultblue}{#1}}
\newcommand{\oursone}[1]{\textcolor{resultorange}{#1}}

\begin{tabularx}{\linewidth}{
@{}>{\raggedright\arraybackslash}p{.245\linewidth}
@{\hspace{2.8pt}}NNNHZZS@{}
}
\toprule
&
\multicolumn{3}{c}
{\textbf{VLA / WAM}} &
\multicolumn{1}{>{\columncolor{hybridfill}}c}{\textbf{Hybrid}} &
\multicolumn{2}{>{\columncolor{zeroshotfill}}c}
{\textbf{0-shot}} &
\multicolumn{1}{>{\columncolor{oneshotfill}[\tabcolsep][\tabcolsep]}c}
{\textbf{1-shot}} \\

\textbf{Task} &
\resulthead{Galaxea}{G0.5} &
\raisebox{.5\baselineskip}[0pt][0pt]{\mbox{OpenWAM $\alpha$}} &
\raisebox{.5\baselineskip}[0pt][0pt]{$\pi_{0.5}$} &
\resulthead{\gptastra{}}{$+\pi_{0.5}$} &
\resulthead{\gptastra{}}{Direct} &
\ourszero{\resulthead{\gptastra{}}{\textbf{RoboICL}}} &
\oursone{\resulthead{\gptastra{}}{\textbf{RoboICL}}} \\
\midrule

Organize the table
& 46.33 & 62.50 & 23.33 & 60.00 & 30.00
& \ourszero{45.00} & \oursone{50.00} \\

Classify by language
& 1.07 & 1.33 & 0.60 & 38.00 & 60.00
& \ourszero{44.00} & \oursone{$44.00^\dagger$} \\

Imitate sorting sequence
& 1.67 & 2.90 & 1.60 & 53.00 & 0.00
& \ourszero{90.00} & \oursone{90.00} \\

Arrange largest number
& 4.11 & 4.36 & 2.29 & 50.00 & 57.00
& \ourszero{71.00} & \oursone{58.00} \\

Pack objects into a box
& 17.12 & 20.83 & 18.36 & 50.00 & 50.00
& \ourszero{16.00} & \oursone{36.00} \\

Classify objects
& 10.33 & 5.53 & 24.67 & 71.00 & 100.00
& \ourszero{100.00} & \oursone{71.00} \\

Build Tower
& 82.93 & 52.53 & 37.73 & 64.00 & 12.00
& \ourszero{16.00} & \oursone{100.00} \\

Make a Kong in Mahjong
& 90.00 & 32.00 & 26.67 & 40.00 & 0.00
& \ourszero{0.00} & \oursone{0.00} \\

Fold clothes
& 32.75 & 51.31 & 29.12 & 100.00 & 40.00
& \ourszero{4.00} & \oursone{84.00} \\

Put bottles in a bin
& 96.30 & 94.03 & 79.93 & 100.00 & 36.00
& \ourszero{70.00} & \oursone{73.00} \\

\midrule
\textbf{Overall}
& \textbf{38.26}
& {32.73}
& {24.43}
& \textbf{62.60}
& {38.50}
& {\ourszero{45.60}}
& \oursone{\textbf{60.60}} \\
\bottomrule
\end{tabularx}

\endgroup
\end{table}

\paragraph{RoboDojo leaderboard comparison.}
Figure~\ref{fig:teaser} and Table~\ref{tab:category-comparison} compare RoboICL with official zero-shot \gptastra{} and four published robot-policy baselines across all 30 tasks in Open, Memory, Precision, and Long-Horizon, using RoboDojo's public layouts and native scoring. Each category equally averages its complete task set. Tasks with substantial layout variation receive 50 evaluations, while stable tasks receive 20; tasks whose evaluation is discontinued are waived and assigned zero. Table~\ref{tab:category-comparison} reports per-task evaluation counts.

RoboICL leads the compared category means on Memory and Open. Its Precision score of 38.53 is comparable to Liber-0 Preview's 38.28. It ranks third on Long-Horizon, behind Simate-beta and Liber-0 Preview, and achieves the highest 30-task Overall score at 50.64.

\begin{table}[t]
\centering
\caption{\textbf{Four-category RoboDojo comparison.} Mean progress score (0--100). Published VLA/WAM columns use three seeds and 50 episodes per task; official zero-shot \gptastra{} (RoboProbe) uses one seed and 50 episodes per task~\citep{chen2026robodojo,robodojo2026astraeval}. All RoboICL evaluations use seed 0. RoboICL uses zero shot on Open and one shot elsewhere; blue and orange cells mark these settings. Upper-right markers give the number of evaluated layouts; 0 marks a waived task assigned zero. Bold marks the highest score in each row.}
\label{tab:category-comparison}
\begingroup
\scriptsize
\setlength{\tabcolsep}{2.2pt}
\renewcommand{\arraystretch}{1.08}
\setlength{\aboverulesep}{0pt}
\setlength{\belowrulesep}{0pt}
\arrayrulecolor{resultrule}
\newcolumntype{Q}{>{\centering\arraybackslash}X}
\newcommand{\zeroshotcell}[1]{\cellcolor{zeroshotfill}#1}
\newcommand{\oneshotcell}[1]{\cellcolor{oneshotfill}#1}
\newcommand{\layoutscore}[2]{\makebox[\linewidth][c]{#1}\llap{\raisebox{-.35ex}[0pt][0pt]{$\,^{#2}$}}}
\newcommand{\compactcite}[1]{\mbox{\scalebox{0.78}{\tiny\citep{#1}}}}
\newcommand{\categoryhead}[1]{\hline\multicolumn{7}{c}{\textit{#1}}\\\hline}
\begin{tabularx}{\linewidth}{@{}>{\raggedright\arraybackslash}p{.245\linewidth}@{\hspace{2.8pt}}QQQQQQ@{}}
\toprule
& \multicolumn{4}{c}{\textbf{VLA / WAM}} & \multicolumn{2}{c}{\textbf{\gptastra{}}}\\
& Galaxea G0.5 & DM0.5 & Liber-0 Preview & Simate-beta & RoboProbe & \\
\multirow{-2}{*}{\textbf{Task}} & \compactcite{liu2026g0} & \compactcite{dm05} & \compactcite{robodojo2026astraeval} & \compactcite{robodojo2026astraeval} & \compactcite{robodojo2026astraeval} & \multirow{-2}{*}{\textbf{RoboICL}}\\
\categoryhead{Open}
Align Blocks & 0.00 & 0.00 & 0.00 & 0.00 & \zeroshotcell{50.00} & \zeroshotcell{\layoutscore{\textbf{90.00}}{50}} \\
Classify Objects by Language & 1.07 & 0.47 & 3.87 & 6.00 & \zeroshotcell{46.00} & \zeroshotcell{\layoutscore{\textbf{70.80}}{50}} \\
General Pickup & 12.67 & 14.00 & 34.00 & 49.33 & \zeroshotcell{84.00} & \zeroshotcell{\layoutscore{\textbf{90.00}}{50}} \\
Pick from Conveyor by Image & 0.00 & 0.00 & 0.00 & 0.00 & \zeroshotcell{4.00} & \zeroshotcell{\layoutscore{\textbf{22.00}}{50}} \\
Pour by Language & 0.00 & 0.00 & 0.00 & 0.00 & \zeroshotcell{\textbf{0.40}} & \zeroshotcell{\layoutscore{0.00}{0}} \\
Solve Equation & 0.00 & 0.00 & 0.00 & 0.00 & \zeroshotcell{40.00} & \zeroshotcell{\layoutscore{\textbf{86.00}}{50}} \\
Stack Blocks by Language & 0.13 & 4.80 & 15.60 & 17.33 & \zeroshotcell{48.00} & \zeroshotcell{\layoutscore{\textbf{79.20}}{50}} \\
Store Tools in Toolbox & 0.00 & 0.17 & 0.00 & 0.33 & \zeroshotcell{\textbf{2.50}} & \zeroshotcell{\layoutscore{0.00}{0}} \\
\textbf{Category mean} & 1.73 & 2.43 & 6.68 & 9.12 & \zeroshotcell{34.36} & \zeroshotcell{\textbf{54.75}} \\
\categoryhead{Memory}
Cover Blocks & 20.67 & \textbf{100.00} & \textbf{100.00} & \textbf{100.00} & \zeroshotcell{49.30} & \oneshotcell{\layoutscore{\textbf{100.00}}{20}} \\
Imitate Sorting Sequence & 1.67 & 1.80 & 8.63 & 4.67 & \zeroshotcell{58.90} & \oneshotcell{\layoutscore{\textbf{75.00}}{20}} \\
Match \& Pick from Conveyor & 29.33 & \textbf{70.67} & 64.00 & 56.00 & \zeroshotcell{62.00} & \oneshotcell{\layoutscore{65.00}{20}} \\
Press by Number & 0.00 & 95.33 & 0.00 & 0.67 & \zeroshotcell{70.00} & \oneshotcell{\layoutscore{\textbf{100.00}}{20}} \\
Swap T & 0.00 & 0.00 & 53.33 & 0.00 & \zeroshotcell{18.00} & \oneshotcell{\layoutscore{\textbf{80.00}}{20}} \\
Swap Blocks & 0.00 & 18.67 & 0.67 & \textbf{38.67} & \zeroshotcell{0.00} & \oneshotcell{\layoutscore{0.00}{0}} \\
\textbf{Category mean} & 8.61 & 47.74 & 37.77 & 33.33 & \zeroshotcell{43.04} & \oneshotcell{\textbf{70.00}} \\
\categoryhead{Precision}
Build Tower & 82.93 & 55.20 & 84.33 & \textbf{87.40} & \zeroshotcell{16.40} & \oneshotcell{\layoutscore{80.60}{50}} \\
Deposit Coin & 10.93 & 13.87 & 18.67 & 16.40 & \zeroshotcell{16.00} & \oneshotcell{\layoutscore{\textbf{78.00}}{50}} \\
Fasten Screws & 30.00 & 40.20 & \textbf{61.80} & 40.93 & \zeroshotcell{36.40} & \oneshotcell{\layoutscore{49.80}{50}} \\
Insert Key & 14.90 & 4.90 & 0.90 & 10.90 & \zeroshotcell{14.40} & \oneshotcell{\layoutscore{\textbf{15.00}}{20}} \\
Insert Tubes & 58.53 & 71.73 & \textbf{82.53} & 41.20 & \zeroshotcell{6.00} & \oneshotcell{\layoutscore{40.80}{50}} \\
Play Xylophone & 0.00 & 0.67 & 0.00 & 3.33 & \zeroshotcell{8.00} & \oneshotcell{\layoutscore{\textbf{44.00}}{50}} \\
Plug in Charger & 0.67 & 2.67 & 27.33 & \textbf{29.33} & \zeroshotcell{0.00} & \oneshotcell{\layoutscore{0.00}{0}} \\
Pour Balls into Vase & 28.00 & 9.33 & 30.67 & \textbf{45.33} & \zeroshotcell{4.00} & \oneshotcell{\layoutscore{0.00}{0}} \\
\textbf{Category mean} & 28.25 & 24.82 & 38.28 & 34.35 & \zeroshotcell{12.65} & \oneshotcell{\textbf{38.53}} \\
\categoryhead{Long-Horizon}
Classify Objects & 10.33 & 26.83 & 19.73 & 61.63 & \zeroshotcell{69.00} & \oneshotcell{\layoutscore{\textbf{85.50}}{50}} \\
Fill Egg Holder & 3.03 & 1.37 & 12.57 & 16.60 & \zeroshotcell{5.50} & \oneshotcell{\layoutscore{\textbf{21.30}}{50}} \\
Fill Pen Holder & 41.27 & 22.10 & 49.77 & \textbf{50.37} & \zeroshotcell{5.00} & \oneshotcell{\layoutscore{29.80}{50}} \\
Make a Kong in Mahjong & \textbf{90.00} & 56.67 & 18.00 & 76.00 & \zeroshotcell{0.00} & \oneshotcell{\layoutscore{0.00}{0}} \\
Organize the Table & 46.33 & 44.00 & 47.50 & \textbf{71.83} & \zeroshotcell{39.50} & \oneshotcell{\layoutscore{55.00}{20}} \\
Play Stacking Toy & 0.47 & 1.20 & \textbf{26.73} & 0.00 & \zeroshotcell{6.60} & \oneshotcell{\layoutscore{0.00}{0}} \\
Play Tic-Tac-Toe & 65.23 & 35.77 & 95.67 & 86.27 & \zeroshotcell{10.70} & \oneshotcell{\layoutscore{\textbf{98.75}}{20}} \\
Put Bottles into Dustbin & 96.30 & 81.70 & 97.90 & \textbf{100.00} & \zeroshotcell{35.30} & \oneshotcell{\layoutscore{62.50}{20}} \\
\textbf{Category mean} & 44.12 & 33.70 & 45.98 & \textbf{57.84} & \zeroshotcell{21.45} & \oneshotcell{44.11} \\
\hline\textbf{Overall} & 21.48 & 25.80 & 31.81 & 33.68 & 26.86 & \textbf{50.64} \\
\bottomrule
\end{tabularx}
\endgroup
\end{table}

\paragraph{Gains on memory and precision tasks.}
RoboICL combines leading Memory and Open scores with competitive Precision and Long-Horizon performance; its 50.64 Overall score in Table~\ref{tab:category-comparison} is 16.96 points above the strongest comparison method. The task rows show where the advantage arises. Anchor zero preserves the episode's first observation, retaining information needed when the goal depends on the initial state. On two such tasks, Cover Blocks and Swap T, RoboICL scores 100 and 80, compared with 49.30 and 18.00 for official zero-shot \gptastra{}, respectively. The gains also extend beyond remembering the initial scene. Imitate Sorting Sequence reaches 75.00 against 8.63 for the strongest VLA/WAM baseline, while the precision tasks Deposit Coin and Play Xylophone improve the corresponding strongest baselines by 59.33 and 40.67 points.

\paragraph{Grasp and insertion correction on Deposit Coin.}
Deposit Coin requires more than lifting a small, thin object: a score of 100 requires the coin's bounding box to enter the bank's narrow slot region and both arms to return to origin. Every comparison method in Table~\ref{tab:category-comparison} scores at most 18.67, whereas RoboICL averages 78.00 over 50 layouts, including 37 full-score episodes. The selected layout-0 episode exposes the closed-loop precision behind this aggregate result. At step 15, \gptastra{} explicitly switches from coarse approach to ``\emph{wrist imagery for millimetric centering.}'' It detects failed retention from RGB at steps 45, 65, and 90, changes both overlap and grasp height, confirms a stable pinch at step 110, and then transfers the coin between the two hands.

\begin{figure}[t]
  \centering
  \includegraphics[width=\linewidth]{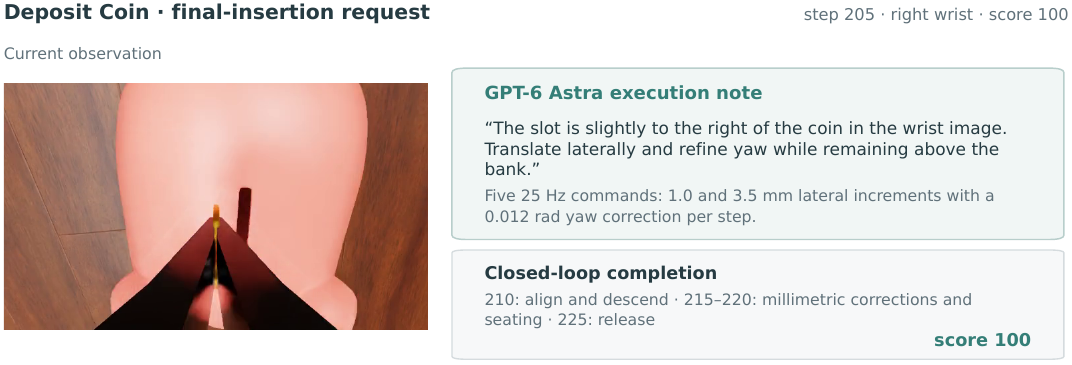}
  \vspace{-6pt}
  \caption{\textbf{The critical wrist-guided request in Deposit Coin.} The figure shows the right-wrist observation from the request at step 205. \gptastra{} identifies the coin--slot offset and emits five 25\,Hz commands with millimeter-scale translation and a small yaw correction. Subsequent requests complete alignment, seating, and release; the selected episode receives score 100.}
  \label{fig:coin-recovery}
\end{figure}

The decisive insertion request occurs at step 205 (Figure~\ref{fig:coin-recovery}). From the current right-wrist image, the model identifies which side of the slot remains misaligned and emits five small Cartesian and yaw corrections while holding the coin above the bank. The next requests align and descend, correct the remaining offset by a few millimeters, seat the coin, and release it at step 225. The wrist view confirms release at step 230 with score 100 at step 259. This trace shows how the demonstration supplies the task procedure while interaction memory and the latest observation support repeated visual correction within that procedure. Appendix~\ref{app:additional-cases} provides complementary examples on contact-rich screw fastening and bimanual role adaptation in Fill Pen Holder.

\begin{figure}[!ht]
  \centering
  \includegraphics[width=\linewidth]{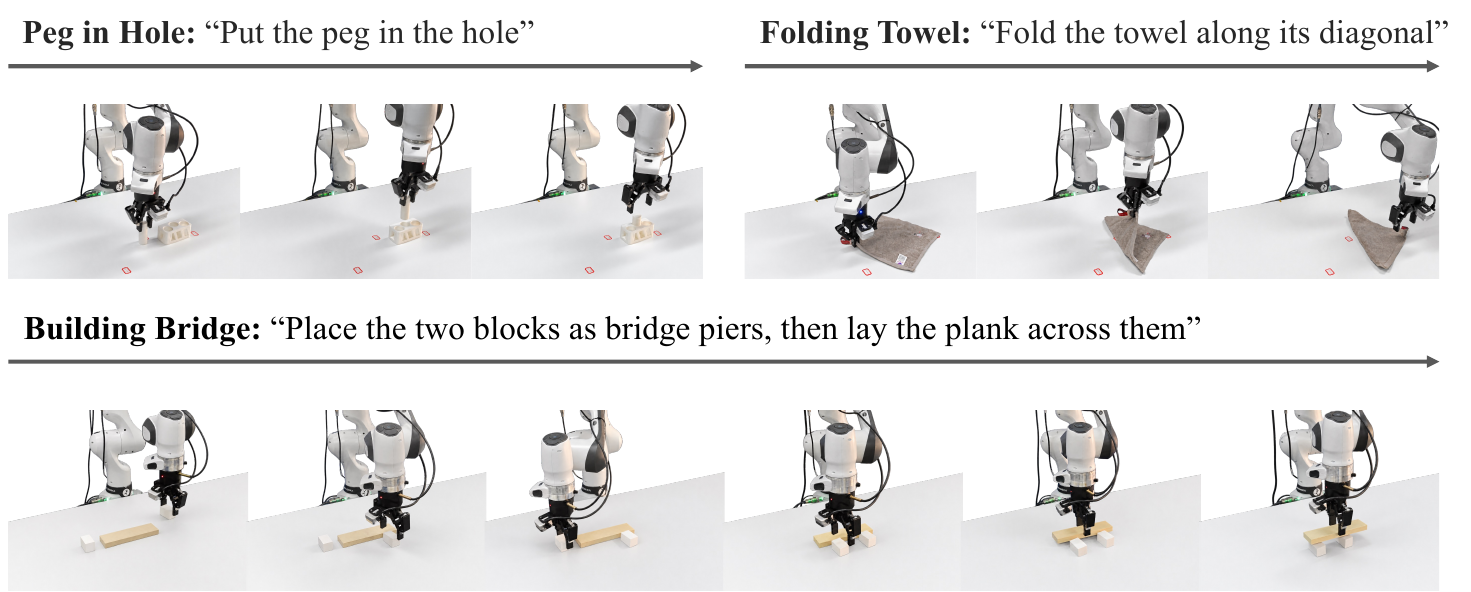}
  \vspace{-10pt}
  \caption{\textbf{Real-robot manipulation.} Executions of Peg in Hole, Folding Towel, and Building Bridge using a Franka Research 3 with wrist and external RGB cameras.}
  \label{fig:real_robot}
\end{figure}

\subsection{Transfer to real-robot manipulation}
\label{sec:realrobot}
To test demonstration-conditioned control on a different embodiment, we evaluate RoboICL on a Franka Research 3 arm. A human expert collects three demonstrations per task through GELLO, and observations combine a wrist-mounted D405 camera with an external D515 camera. We test \texttt{Peg in Hole}, \texttt{Folding Towel}, and \texttt{Building Bridge}, spanning precision, deformable-object, and long-horizon manipulation. Each zero-, one-, and three-shot condition contains five trials with randomized object positions and orientations. The metric averages normalized credit for predefined task stages. Figure~\ref{fig:real_robot} illustrates the three tasks.

\begin{figure}[t]
  \centering
  \includegraphics[width=0.8\linewidth]{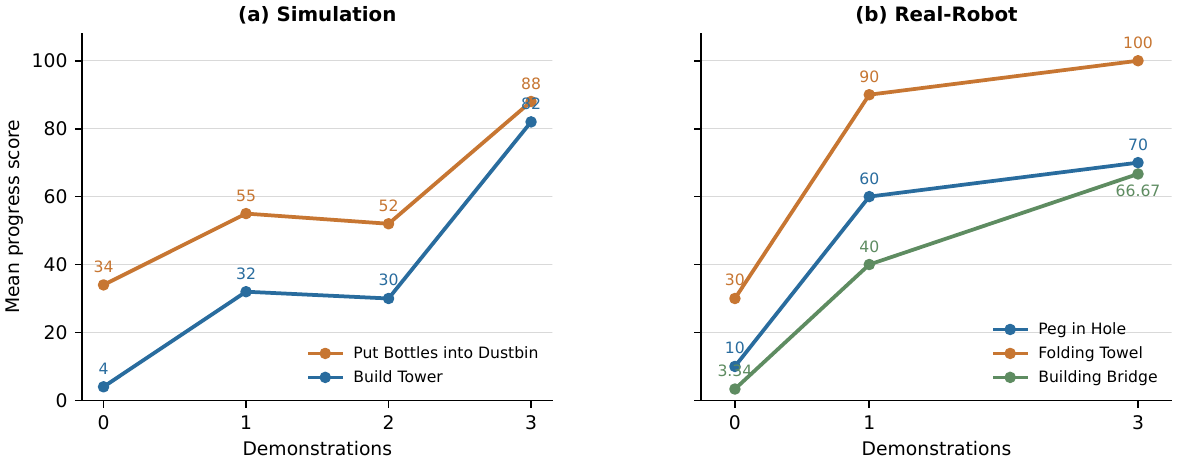}
  \caption{\textbf{Shot scaling in simulation and on a real robot.} Left: five-layout simulator means at $H=15$ and $J=B=5$. Right: five-trial real-robot means under staged progress scoring. Three demonstrations improve every shown task over zero shot.}
  \label{fig:demonstration-scaling}
\end{figure}

On the real robot, every task improves from zero to one to three demonstrations (Figure~\ref{fig:demonstration-scaling}, right); the three-task mean rises from 14.45 to 63.33 and 78.89. This result extends demonstration conditioning to precision, deformable-object, and longer-horizon real-robot manipulation without parameter updates. At three shots, two of five \texttt{Peg in Hole} trials complete insertion, and \texttt{Building Bridge} gains 26.67 points over one shot.

The left panel supplies a complementary simulator shot-scaling study at $H=15$ and $J=B=5$. Both tasks improve from zero to three demonstrations, with a dip from one to two shots. Every added demonstration contributes five blocks, whereas the context-allocation study holds $J+B$ fixed.

\begin{wraptable}{r}{.44\linewidth}
\centering
\vspace{-10pt}
\caption{\textbf{Folding Towel OOD progress scores.} Three-shot mean progress score (0--100) over five trials per towel. Unseen towels do not appear in the demonstrations.}
\label{tab:towel-ood}
\footnotesize
\setlength{\tabcolsep}{5pt}
\begin{tabular}{@{}lr@{}}
\toprule
Test condition & Progress score \\
\midrule
Seen 1                  & 100.00 \\
Unseen 1                &  90.00 \\
Unseen 2 (larger towel) &  40.00 \\
\midrule
Mean across unseen conditions & 65.00 \\
\bottomrule
\end{tabular}
\vspace{-10pt}
\end{wraptable}

To examine robustness beyond the demonstrated instance, we further evaluate \texttt{Folding Towel} in the three-shot setting with two unseen towels. Performance remains high on Unseen 1, whereas the larger Unseen 2 scores 40, which is 60 points below the seen towel, and only one of its five trials receives full credit (Table~\ref{tab:towel-ood}). Although the high-level goal remains a diagonal fold, the model does not reliably adapt its motion to the change in object scale. This behavior suggests that the frozen model uses demonstrations to ground broad folding knowledge in the embodiment and action space but may follow the demonstrated motion too closely. Appendix~\ref{app:realrobot-details} describes the detailed failure modes and OOD examples.

\subsection{Inference cost, latency, and cache reuse}
\label{sec:efficiency}

Having established control performance, we measure its inference demand under the context and action interfaces of Section~\ref{sec:efficient-interface}: cached tokens, model calls, API latency, and end-to-end wall time.

\paragraph{Token use and elapsed time.}
Table~\ref{tab:resource-accounting} uses three tasks for which all five zero-shot and five one-shot episodes from Table~\ref{tab:ten-task} have complete usage records, totaling 30 episodes. Each request uses native $1920\times480$ JPEG triptychs, comprising three $640\times480$ views, with \texttt{original} image detail and a 50-image limit. The zero-shot setting uses $B=25$; one shot uses $J=B=12$.

\begin{table}[t]
\centering
\caption{\textbf{Inference use and latency on three five-layout task pairs.} Each row averages five scored episodes. Tokens are provider-reported input plus output; KV-cache hit rate is the token-weighted cached share of input. Calls count completed API requests, chunks count executed tool calls, API time sums request latency, and wall time includes simulator and transport overhead.}
\label{tab:resource-accounting}
\footnotesize
\setlength{\tabcolsep}{3pt}
\begin{tabularx}{\linewidth}{@{}Xrrrrrrr@{}}
\toprule
Task & Shot & \shortstack{Progress\\score} & \shortstack{Tokens\\(M)} & \shortstack{KV-cache\\hit (\%)} & \shortstack{Calls /\\chunks} & \shortstack{API\\(min)} & \shortstack{Wall\\(min)} \\
\midrule
Build Tower & 0 & 16 & 5.040 & 91.66 & 77.4 / 72.4 & 48.8 & 58.7 \\
& 1 & 100 & 3.265 & 92.28 & 40.2 / 39.6 & 31.5 & 37.9 \\
\midrule
Classify Objects & 0 & 100 & 2.632 & 89.27 & 55.6 / 53.4 & 29.9 & 37.4 \\
& 1 & 71 & 5.573 & 94.03 & 62.6 / 62.2 & 43.3 & 52.8 \\
\midrule
Put Bottles into Dustbin & 0 & 70 & 3.246 & 90.93 & 64.4 / 61.2 & 31.4 & 37.9 \\
& 1 & 73 & 5.091 & 93.78 & 61.8 / 61.0 & 38.8 & 46.3 \\
\bottomrule
\end{tabularx}
\end{table}

Demonstrations can shorten an episode while enlarging each request. On Build Tower, mean progress score increases from 16 to 100 as calls fall from 77.4 to 40.2 and total tokens from 5.04 to 3.27 million. Classify Objects and Put Bottles consume more tokens with one shot. Across the six conditions, mean uncached input ranges from 248,000 to 413,000 tokens per episode and mean output from 45,000 to 85,000.

Across all 1,810 completed requests, the provider reports 113.15 million of 122.51 million input tokens as cached, a token-weighted KV-cache hit rate of 92.4\%. The remaining 9.35 million input tokens are uncached, and total input plus output is 124.24 million tokens.

\begin{figure}[!htbp]
  \centering
  \includegraphics[width=\linewidth]{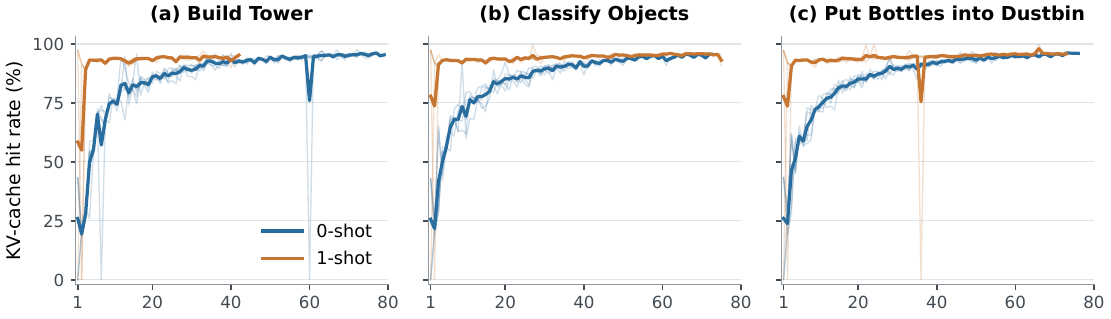}
  \vspace{-10pt}
  \caption[KV-cache hit rate across model calls.]{\textbf{KV-cache hit rate across model calls.} The same three-task, 30-episode panel as Table~\ref{tab:resource-accounting}. Thin lines show each layout; thick lines give the pooled token-level rate $100\sum_e C_{e,t}/\sum_e I_{e,t}$ at each call index. Later points average the episodes that reach that call. Curves use completed responses with provider usage and include all input modalities.}
  \label{fig:cache-reuse}
\end{figure}

\paragraph{KV-cache hit rate across model calls.}
Figure~\ref{fig:cache-reuse} resolves the episode totals into individual model calls. For episode $e$ and completed-call index $t$, the token-level KV-cache hit rate is $\rho_{e,t}=C_{e,t}/I_{e,t}$, where $C$ and $I$ are provider-reported cached and total input tokens. The traces reach high hit rates after the initial calls, with occasional sharp drops hidden by episode averages. Pooling input tokens across the three tasks, the first-call rate is 25.9\% at zero shot and 71.1\% at one shot; over calls six onward, it reaches 91.4\% and 94.3\%, respectively.

\paragraph{Relation to external resource reports.}
In a public resource report, GPT-as-Policy averages 22.65 million tokens per episode for direct \gptastra{} and 12.50 million for its $\pi_{0.5}$ hybrid over 50 episodes per method~\citep{gptaspolicy2026cost}. Their 154.6 and 75.5 reported decisions are executed action chunks, whereas Table~\ref{tab:resource-accounting} separately counts API requests and executed chunks. Their token-level KV-cache hit rates are approximately 97--98\%, with mean uncached input of approximately 458,000 and 318,000 tokens per episode. These figures provide a resource reference under their published protocol; Table~\ref{tab:resource-accounting} provides the corresponding decomposition for RoboICL.

\FloatBarrier
\subsection{Reducing model calls with Jev}
\label{sec:jev}
Cache reuse reduces repeated input processing, while every fresh model call still incurs inference. Jev provides a complementary mechanism: execute more of an existing proposal when feedback indicates that replanning is unnecessary. We evaluate it as an optional gate after the main RoboICL benchmarks. We tested Jev, a System One decision model that returns typed, probability-backed judgments from structured state and questions~\citep{jevai2026docs}, as a gate for reusing a \gptastra{}-predicted action suffix. The zero-shot simulator study covers standard layouts 0--4 of \emph{Align Blocks} and \emph{General Pickup}, with five episodes per task and condition. Both conditions use \gptastra{} with \texttt{xhigh} reasoning, the same task prompt, cameras, action constraints, 20 predicted actions per request, up to five executed actions per segment, and 25 interaction-memory slots. Pure \gptastra{} replans after each segment. Jev can approve one further five-action segment from the current plan using proprioception, execution tracking, and the plan note, but no RGB. It therefore controls action reuse without generating actions. The gate configuration was selected on these layouts, making this a development-set evaluation.

\begin{table}[!ht]
\centering
\caption{\textbf{Complete-task success and paired completion time with Jev.} Counts are over five layouts. The last column sums wall-clock seconds only over layouts successful in both conditions, in the order pure \gptastra{}/\gptastra{}+Jev; $n$ gives the number of jointly successful layouts.}
\label{tab:jev-main}
\small
\setlength{\tabcolsep}{5pt}
\begin{tabular}{@{}lccc@{}}
\toprule
Task & Pure \gptastra{} & \gptastra{}+Jev & Paired wall time (s) \\
\midrule
Align Blocks & 3/5 & 2/5 & 1128/865 ($n=1$) \\
General Pickup & 4/5 & 5/5 & 3642/2327 ($n=4$) \\
\bottomrule
\end{tabular}
\end{table}

Table~\ref{tab:jev-main} shows the success tradeoff: General Pickup increases from 4/5 to 5/5, while Align Blocks changes from 3/5 to 2/5. On the four jointly successful General Pickup layouts, wall time falls by 36.1\%. Across all five layouts, \gptastra{} requests fall from 181 to 121 on Align Blocks and from 116 to 60 on General Pickup; provider-reported \gptastra{} tokens fall from 9.53 to 5.08 million and from 4.95 to 1.39 million, respectively. Jev adds approximately 0.86 seconds per gate call.

On \emph{General Pickup} layout 3, Jev uses 103 robot steps versus pure \gptastra{}'s 75 despite 13 versus 15 requests: fewer model calls can coexist with more robot actions. Appendix~\ref{app:jev}, including Tables~\ref{tab:jev-layouts} and~\ref{tab:jev-runtime}, reports every outcome and runtime measurement. The study establishes substantial \gptastra{} call and token reductions on both tasks, with task-dependent effects on success.

\FloatBarrier
\section{Conclusion and limitations}
\label{sec:discussion}
\label{sec:conclusion}
RoboICL organizes demonstrations and online experience into a shared context and control interface for a frozen multimodal model. Execution receipts connect actions to realized outcomes, bounded anchors retain temporally distributed visual evidence. RoboICL leads the compared RoboDojo methods on Memory and Open, achieves comparable performance on Precision, and remains competitive on Long-Horizon. Demonstration scaling on three real-robot tasks extends the approach beyond simulation without parameter updates.

Efficiency depends on both context reuse and feedback frequency. Provider counters show a 92.4\% token-weighted KV-cache hit rate across 1,810 calls. Jev gating reduces \gptastra{} calls and tokens on both evaluated tasks with task-dependent effects on success. The experiments cover \gptastra{}, RoboDojo, and three real-robot tasks; Jev is evaluated on its development layouts. Broader models and hardware platforms, and held-out Jev evaluation are the next tests of generality.

\label{main-text-end}

\clearpage
\bibliography{references}
\bibliographystyle{roboicl}
\clearpage
\appendix
\section{Method and implementation}

\subsection{Context construction}
\label{app:context-construction}

\paragraph{Interaction memory.}
Interaction memory exposes how earlier decisions change the environment. After decision round $t$, the execution system returns receipt $r_t$ and the next observation $(o_{t+1},z_{t+1})$, thereby completing interaction $c_t$. The deterministic update operator inserts this interaction and applies the retention policy in Section~\ref{sec:memory}, allowing a later decision to relate the current state to earlier actions, task progress, unsuccessful contacts, and execution interruptions.

Retaining every completed interaction would make the request grow with episode length. RoboICL instead allocates $B$ slots to interaction memory. The first $B-1$ slots provide fixed temporal anchors across the episode, while the final slot preserves the latest completed interaction. With $T$ denoting the maximum number of environment steps and $i\in\{0,\ldots,B-2\}$ indexing a fixed anchor, its target is $\tau_i=\lfloor iT/B\rfloor$. The initial nonempty interaction occupies anchor zero, and every later anchor stores the first completed interaction $c_j$ whose interval satisfies $s_j<\tau_i\le e_j$. The latest-interaction slot supplies the most recent outcome for local correction and is deduplicated if the same interaction occupies an anchor.

Sparse retention can otherwise make observations separated by many environment steps appear adjacent. Explicit \texttt{<TRAJECTORY\_GAP>} records therefore mark intervals between retained interactions, omitting full images and action arrays while preserving interval boundaries and compact execution metadata. For $T=700$ and $B=5$, targets are $0,140,280,420$. The released request retains intervals $[0,15]$, $[133,148]$, $[268,283]$, $[418,433]$, and the latest interval $[672,687]$. At most $B$ complete interactions are retained, requiring at most $2B$ endpoint triptychs before deduplication.

\paragraph{Demonstration context.}
Expert demonstrations supply task knowledge before execution begins. The $K$ demonstrations in $\mathcal D_K$ remain unchanged throughout the episode while $\mathcal M_t$ evolves. Each selected expert interaction contains the recorded action chunk, its endpoint observations, and a synthetic receipt constructed for serialization. Serializing complete demonstrations would consume the multimodal context budget before online interaction begins. RoboICL selects $J$ non-overlapping action chunks from each trajectory. Each selected block preserves the source actions and its start and result observations, while explicit gaps identify omitted intervals. Retained block endpoints contribute at most $2JK+2B$ triptychs before deduplication; The three-shot scaling configuration uses $K=3$, $J=5$, 15 controller steps per block, and $B=5$, yielding at most 40 endpoint triptychs. The main one-shot configuration and its terminal-observation exceptions are specified in Section~\ref{sec:ablation}.

\paragraph{Selecting main demonstration windows.}
The main preparation rule uses approximately uniform targets with deterministic action-validity adjustment. For a trajectory containing $N$ source observations and action horizon $H$, the intended final start is $\ell=N-1-H$. The first and last windows must be valid. We form twelve targets $q_i=\operatorname{round}(i\ell/11)$ for $i=0,\ldots,11$, using nearest rounding with ties to even. Interior starts are selected in temporal order: choose the valid integer start closest to $q_i$ within $[s_{i-1}+H,\ell-H(11-i)]$, breaking distance ties toward the earlier frame. The interval enforces nonoverlap and leaves space for the remaining blocks. Candidates are filtered by the translation and rotation-vector norm bounds for both arms. Subsequent reference validation checks the full action contract and observation alignment. Source actions are preserved without clipping or rewriting.

A terminal-fallback variant chooses the latest valid complete block at or before $N-1-H$ and places the temporal targets relative to that start. The true terminal observation is retained separately if the last selected block ends earlier. This variant applies to Fill Pen Holder, Fill Egg Holder, Play Tic-Tac-Toe, and Play Stacking Toy. Only the latter two leave a terminal gap (15 and 18 source frames, respectively), each requiring one extra triptych. Explicit gaps separate these terminal frames from the final selected block's actual result.

The released evaluation configuration records each demonstration's source episode and selected starts; every result endpoint is start plus $H$. The five-block shot-scaling study uses a gripper-event heuristic followed by endpoint alignment.

\subsection{Concurrent-work interface details}
\label{app:concurrent-interfaces}

GPT-as-Policy Direct~\citep{su2026astra} uses a persistent conversation, history files, writable notes, and image tools to support decisions about the absolute targets in Figure~\ref{fig:zero-shot-protocols}. The distinct GPT-Policy framework~\citep{cheng2026gptpolicy} uses \texttt{move\_to} for an absolute tool-center-point (TCP) pose, \texttt{move\_eef\_chunk} for an ordered sequence of absolute TCP waypoints, and \texttt{set\_gripper} for a separate gripper change. Its controller interpolates waypoints, solves inverse kinematics, and assigns timing, holding gripper targets fixed during each motion request. Thus it supports numerical action chunks, but the controller determines the intervening samples and timing. RoboICL's single \act{} tool instead accepts an $H\times d_a$ matrix specifying Cartesian increments and gripper targets at every step, executed at 25\,Hz in RoboDojo.

Both concurrent frameworks return execution feedback. GPT-Policy limits older live images while retaining accumulated text or host-generated summaries; RoboICL retains anchored visual endpoints and the latest interaction under a shared demonstration/memory budget, uses the same observation--action--receipt--observation grammar for both sources, and marks executed prefixes and omitted intervals through receipts and \texttt{<TRAJECTORY\_GAP>} records. GPT-Policy evaluates ten real-robot tasks with three trials per condition; its evaluation supports the interface comparison here, while our benchmark comparisons use methods evaluated on RoboDojo.

\FloatBarrier
\section{Evaluation protocols and analyses}

\subsection{Additional qualitative case studies}
\label{app:additional-cases}
\begin{figure}[!ht]
  \centering
  \includegraphics[width=\linewidth]{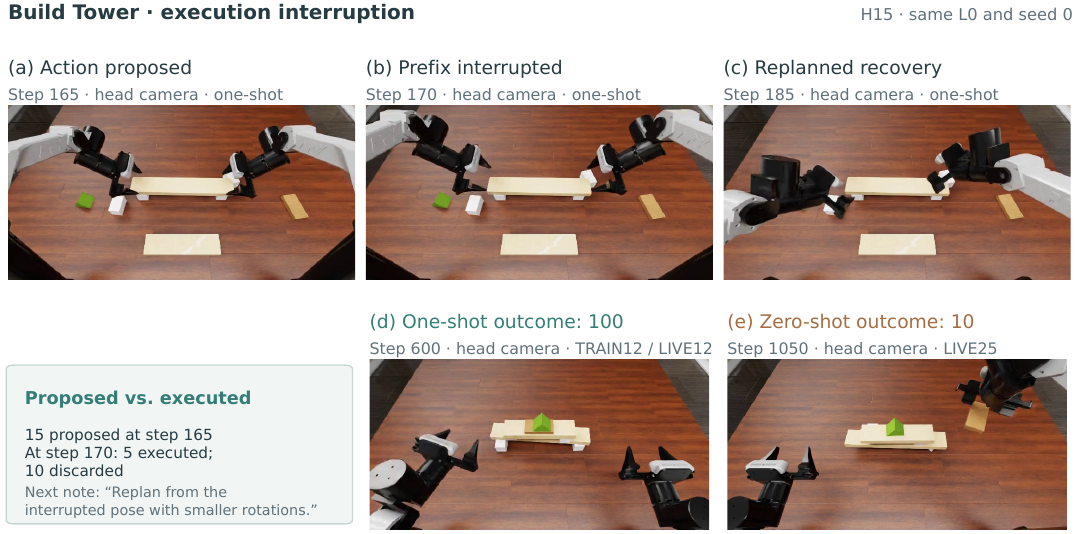}
  \caption{\textbf{Recovery after an interrupted action prefix.} On Build Tower L0, the one-shot policy ($H=15$, $J=B=12$) proposes 15 commands at step 165. The controller executes five before a continuity rejection at step 170 and discards the ten-command suffix. The next execution note requests smaller rotations; construction continues and reaches score 100 at step 600. The same layout and seed under zero-shot RoboICL ($B=25$) receive 10 at the 1050-step limit. All panels use native head-camera frames.}
  \label{fig:tower-prefix}
\end{figure}

\paragraph{Replanning after an interrupted prefix.}
Figure~\ref{fig:tower-prefix} connects the method's proposed--executed distinction to a recorded Build Tower trajectory. At step 165, the policy proposes 15 commands, but the controller executes only five before rejecting the next command. The following execution note requests smaller rotations, and the rollout subsequently completes the tower. A zero-shot rollout on the same layout receives 10 at the step limit.

\FloatBarrier

\begin{figure}[t]
  \centering
  \includegraphics[width=\linewidth]{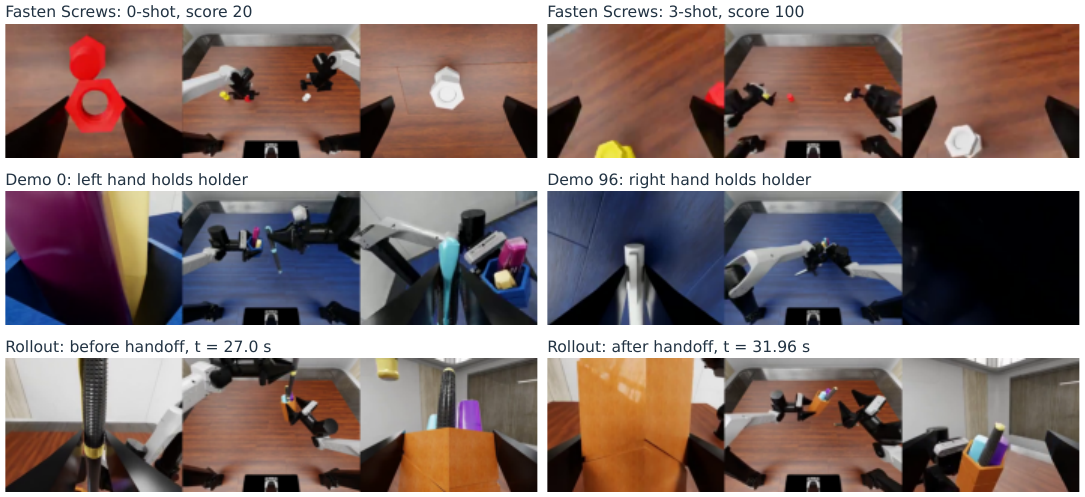}
  \caption{\textbf{Complementary examples of demonstration-conditioned manipulation.} Top: terminal views from Fasten Screws layout 2, where zero-shot RoboICL scores 20 and three-shot RoboICL scores 100. Middle: Fill Pen Holder demonstrations with opposite hand assignments. Bottom: the selected rollout before and after transferring the filled holder from the right hand to the left. Each observation concatenates the left-wrist, head, and right-wrist views in that order.}
  \label{fig:additional-cases}
\end{figure}

\paragraph{Contact strategy in Fasten Screws.}
The selected layout-2 pair contrasts zero-shot control, which scores 20, with a three-shot rollout that scores 100 (Figure~\ref{fig:additional-cases}, top). The three-shot policy brings the yellow and white nuts to their corresponding screws, fastens them through repeated short clockwise motions with release-and-regrasp resets, and completes the red assembly after an additional recovery. The zero-shot rollout instead spends much of its longer execution searching for a stable grasp. The comparison shows how demonstrations can supply both the operation order and the contact strategy for a precision task.

\paragraph{Bimanual role adaptation in Fill Pen Holder.}
The demonstrations expose two valid hand assignments: episode 0 holds the holder with the left hand and inserts with the right, while episodes 96 and 99 reverse those roles. In the selected seed-1/layout-0 rollout, the policy begins with the right hand holding the filled holder, brings it to the left gripper, establishes a new left-hand grasp, and releases the right gripper. The rollout therefore connects role assignments that appear separately in the demonstrations, executing a stable mid-air transfer while the pens remain in the holder. A policy-service failure ends the recording after step 802 before an official task score is returned.

\FloatBarrier

\subsection{Real-robot protocol and failure analysis}
\label{app:realrobot-details}

\paragraph{Platform and demonstrations.}
Experiments use a Franka Research 3 arm. A human expert collects three demonstrations per task using GELLO as the teleoperation interface. Visual observations comprise RGB images from a wrist-mounted D405 camera and an external third-person D515 camera. Zero-shot evaluation provides no demonstration, one-shot evaluation uses the same selected demonstration in every trial, and three-shot evaluation uses all three demonstrations.

\paragraph{Tasks and evaluation protocol.}
The three tasks span high-precision, deformable-object, and long-horizon manipulation. \texttt{Peg in Hole} requires insertion with a nominal peg--hole tolerance of $1\,\mathrm{mm}$. \texttt{Folding Towel} requires folding a towel along its diagonal. \texttt{Building Bridge} requires positioning two bridge piers before placing the deck. Each task and shot setting contains five trials, with object positions randomized within a $10\,\mathrm{cm}\times10\,\mathrm{cm}$ region and orientations randomized about the gravity-aligned axis. Each trial receives a progress score on the same 0--100 display scale used throughout the paper. For \texttt{Peg in Hole}, picking up and inserting the peg each contribute 50 points. For \texttt{Folding Towel}, picking up the towel and completing the fold each contribute 50 points. For \texttt{Building Bridge}, each of the six pickup or placement stages contributes $100/6$ points. Figure~\ref{fig:demonstration-scaling} reports mean progress score, not complete-task success rate.

\paragraph{Failure analysis.}
Eleven of 15 zero-shot trials receive no progress credit, although some rollouts still exhibit basic manipulation; for example, the robot can pick up a towel corner without completing the diagonal fold. Demonstrations increase mean progress score to 63.33 with one shot and 78.89 with three shots. Three of five three-shot \texttt{Peg in Hole} trials stop after pickup: the robot brings the peg near the hole and then stalls during insertion as successive RGB observations change little after contact. \texttt{Building Bridge} remains the lowest-scoring task at every shot count. It gains 26.67 points between one and three shots, compared with 10 points on each other task, and one of five three-shot trials receives full credit. Its longer trajectory is represented by a bounded set of demonstration blocks and interaction-memory anchors.

\begin{figure}[t]
\centering
\includegraphics[width=\linewidth]{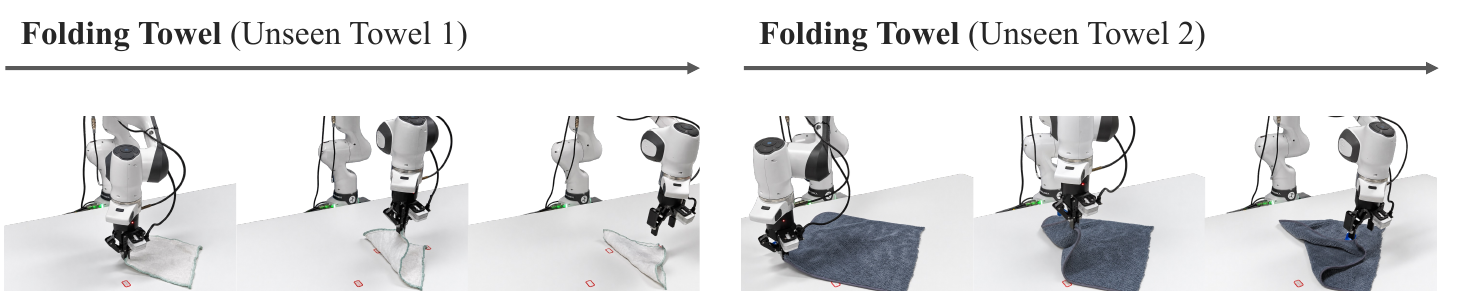}
\caption{Selected three-shot \texttt{Folding Towel} executions with two unseen towels: Unseen 1 (left) and the larger Unseen 2 (right). Table~\ref{tab:towel-ood} reports progress scores over five trials per condition.}
\label{fig:real_robot_ood}
\end{figure}

\FloatBarrier
\section{Cache-prefix checks and detailed Jev results}

\subsection{Cache-prefix verification}
\label{app:cache-verification}
The 30 episodes used for the resource and cache figures contain 1,810 completed requests and 1,780 consecutive-request comparisons. Every marked cache frontier remains byte-identical across its comparison, and each episode retains one fixed-prefix identity. This verifies the stable request prefixes; provider cached-input counters supply the measured KV-cache hit rates.

\subsection{Jev-gated execution: complete five-layout record}
\label{app:jev}
Section~\ref{sec:jev} describes the Jev protocol and reports success counts and paired completion times. Tables~\ref{tab:jev-layouts} and~\ref{tab:jev-runtime} give all 20 scored outcomes and condition-level model use.

\begin{table}[htbp]
\centering
\caption{\textbf{Every Jev-panel layout.} Each cell is progress score (0--100) / executed robot steps / wall-clock seconds (rounded). Scores in this panel are 0 or 100; Failed 200-step episodes are shown but excluded from successful-pair completion-time comparisons.}
\label{tab:jev-layouts}
\small
\setlength{\tabcolsep}{5pt}
\begin{tabular}{@{}llrr@{}}
\toprule
Task & Layout & Pure \gptastra{} & \gptastra{}+Jev \\
\midrule
Align Blocks & 0 & 0/200/4489 & 0/200/1135 \\
& 1 & 100/151/1128 & 100/148/865 \\
& 2 & 100/167/1490 & 0/200/1120 \\
& 3 & 100/138/1054 & 0/200/1199 \\
& 4 & 0/200/2358 & 100/141/827 \\
\midrule
General Pickup & 0 & 100/64/1335 & 100/70/440 \\
& 1 & 100/99/857 & 100/73/557 \\
& 2 & 100/83/792 & 100/84/683 \\
& 3 & 100/75/658 & 100/103/648 \\
& 4 & 0/200/1436 & 100/89/576 \\
\bottomrule
\end{tabular}
\end{table}

\begin{table}[htbp]
\centering
\caption{\textbf{Total model use across each five-layout condition.} \gptastra{} calls and API seconds use completed requests; tokens are provider-reported response usage. Jev reuse counts executed steps approved by the gate. Totals include all five episodes.}
\label{tab:jev-runtime}
\footnotesize
\setlength{\tabcolsep}{4pt}
\begin{tabular}{@{}llrrrrr@{}}
\toprule
Task & Condition & Success & \shortstack{\gptastra{}\\calls} & \shortstack{\gptastra{}\\API s} & \shortstack{\gptastra{}\\tokens} & \shortstack{Jev votes /\\reused steps} \\
\midrule
Align Blocks & Pure \gptastra{} & 3/5 & 181 & 9407 & 9,534,190 & 0 / 0 \\
& \gptastra{}+Jev & 2/5 & 121 & 4101 & 5,083,650 & 113 / 315 \\
\midrule
General Pickup & Pure \gptastra{} & 4/5 & 116 & 4192 & 4,952,887 & 0 / 0 \\
& \gptastra{}+Jev & 5/5 & 60 & 2095 & 1,391,331 & 53 / 140 \\
\bottomrule
\end{tabular}
\end{table}

\subsection{Paired completion time}

Jev and pure \gptastra{} both succeed on \emph{Align Blocks} layout 1 (864.91 versus 1127.98 seconds) and \emph{General Pickup} layouts 0--3 (summed wall times of 2327.17 versus 3641.54 seconds). Jev also succeeds on \emph{Align Blocks} layout 4, where pure \gptastra{} fails, so that episode has no paired completion-time baseline. On \emph{General Pickup} layout 3, Jev takes 103 robot steps versus 75 for pure \gptastra{} while using 13 versus 15 \gptastra{} requests. Fewer model calls can therefore coexist with more physical actions. Failed runs remain in the success denominator and outside the paired completion-time sum.

\end{document}